\documentclass{article}
\usepackage[preprint]{nips_2018}

\usepackage[utf8]{inputenc} 
\usepackage[T1]{fontenc}    
\usepackage{hyperref}       
\usepackage{url}            
\usepackage{booktabs}       
\usepackage{amsfonts}       
\usepackage{nicefrac}       
\usepackage{microtype}      

\usepackage[usenames,dvipsnames,svgnames,table]{xcolor}
\usepackage{amsmath, amsthm, amssymb, bbm, bm, mathtools}
\usepackage{enumerate}
\usepackage{graphicx}
\usepackage{float}
\usepackage{wrapfig, subcaption}
\usepackage[margin=0cm]{caption}
\usepackage[titletoc, toc]{appendix}
\usepackage{tabularx}
\usepackage{booktabs,colortbl}
\usepackage{nicefrac}
\usepackage{blindtext}
\usepackage{marginnote}

\usepackage{multirow}
\usepackage{color}

\DeclareMathOperator*{\argmin}{arg\;min}
\DeclareMathOperator*{\argmax}{arg\;max}
\usepackage{float}
\usepackage{graphics}
\usepackage{graphicx}
\usepackage{algorithm}
\usepackage[noend]{algpseudocode}

\title{Input and Weight Space Smoothing for Semi-supervised Learning}

\author{
  Safa Cicek \\
  University of California, Los Angeles\\
  \texttt{safacicek@ucla.edu} \\
  \And
  Stefano Soatto \\
  University of California, Los Angeles \\
  \texttt{soatto@ucla.edu} \\
}

\begin{document}
\maketitle

\begin{abstract}
We propose regularizing the empirical loss for semi-supervised learning by acting on both the input (data) space, and the weight (parameter) space. We show that the two are not equivalent, and in fact are complementary, one affecting the minimality of the resulting representation, the other insensitivity to nuisance variability. We propose a method to perform such smoothing, which combines known input-space smoothing with a novel weight-space smoothing, based on a min-max (adversarial) optimization. The resulting Adversarial Block Coordinate Descent (ABCD) algorithm performs gradient ascent with a small learning rate for a random subset of the weights, and standard gradient descent on the remaining weights in the same mini-batch. It achieves comparable performance to the state-of-the-art without resorting to heavy data augmentation, using a relatively simple architecture.
\end{abstract}

\section{Introduction}

In semi-supervised learning, we are given $N^l$ labeled samples $x^l \in X^l$ with corresponding labels $y^l \in Y^l$ and $N^u$ unlabeled samples, $x^u \in X^u$. The entire training dataset is $X$ with cardinality $N = N^l + N^u$.  For a discriminative model to exploit unlabeled data, there has to be some prior on the model parameters or on the unknown labels (\cite{chapelle2009semi}). Such a prior can be realized through a regularization functional acting on either the parameters (weight space) or the data (input space).  Both input-space regularization, or ``smoothing'' (\cite{joachims1999transductive,goodfellow2014explaining,miyato2017virtual}) and weight-space smoothing (\cite{hochreiter1997flat,chaudhari2016entropy}) have been shown to improve both supervised (SL) and semi-supervised learning (SSL). The first question we address is whether the two are, in some sense, equivalent.

To answer the question, we conduct experiments that show that, for nonlinear and/or over-parametrized classifiers, input and weight smoothing are not only not equivalent, but they are complementary, suggesting that applying both may be beneficial. The second question we address, therefore, is whether this can be done efficiently and yield performance improvements relative to methods that only smooth one of the two.

To address this, we propose a new algorithm for weight smoothing called \textit{Adversarial Block Coordinate Descent (ABCD)}, which we combine with a standard input-smoothing algorithm (VAT), and test the result on  SSL benchmarks on the CIFAR10 and SVHN benchmark datasets. ABCD combined with VAT achieves state-of-the-art performance with minimal data augmentation (translation and reflection), without complex architectures (e.g. ResNet) and no sophisticated learning machinery.

While in this section our method is motivated heuristically, there are theoretical groundings for performing joint regularization in weight- and input-space, which we discuss in the Supplementary Material. In the next two subsections we describe input and weight smoothing, and in the following subsection we show them to not be equivalent. In the next section we describe the proposed algorithm ABCD, and in the following one we put it to the test on SSL benchmarks.

\subsection{Input smoothing} We call a classifier ``input smooth" when its predictions are robust to small perturbations in the input space. So, input smoothing can be obtained with the following optimization problem: \begin{align}
\min_w  &\  \sum_{x_i \in X} \ell(f(x_i; w), f(x_i+\Delta x_i; w) )  \nonumber \\
{\rm subject \ to} &\ \Delta x_i = \argmax_{||\Delta x_i||<\epsilon_{x}} \ell(f(x_i; w), f(x_i+\Delta x_i; w)) \, \forall x_i \in X
\label{eq:input-smoothing}
\end{align} where $\ell()$ can be cross-entropy, Kullbach-Liebler (KL) divergence or the mean-square error. $f(x; w) \in \mathbb R^K$ is the network output with weights $w$ and $K$ is the number of classes.  This problem can be solved along with minimizing the objective function designed for the task, for instance the cross-entropy loss for classification. In other words, a desirable classifier should not change its predictions for any additive perturbation within a ball of small radius $\epsilon_x$ for any input $x_i$. This idea is also known as max-margin or low-density assumptions in the SSL literature, championed by TSVM (\cite{joachims1999transductive}). Although the perturbations to which we seek insensitivity are unstructured, in imaging data the largest perturbations  are often due to nuisance variability ({\em e.g.}. changes in illumination, vantage point, or visibility).

A popular way of attacking to this min-max problem is through the use of adversarial examples. 
The underlying idea is to add a (regularization) term to the loss function, that penalizes the difference between network outputs for clean samples, and samples with added adversarial noise. Adversarial training (\cite{goodfellow2014explaining}) applies this idea to supervised learning where they change the problem to being robust against noise by moving predictions away from the ground truth labels: \begin{align}
\min_w  &\ \sum_{x_i \in X} \ell(f(x_i; w), f(x_i+\Delta x_i; w) )  \nonumber \\
{\rm subject \ to} &\ \Delta x_i = \argmax_{||\Delta x_i||<\epsilon_x} \ell(P(y_i|x_i), f(x_i+\Delta x_i; w)) \, \forall x_i \in X
\end{align} For this supervised setting, ground truth labels $P(y|x)$ can be used in calculating the adversarial noise $\Delta x$. Instead of finding the exact $\Delta x$ for each input $x$, \cite{goodfellow2014explaining} calculates the first order approximation of adversarial perturbation leading to maximum change in the classifier predictions $f(x;w)$: \begin{align}
& \Delta x \approx \epsilon_x \frac{g}{||g||_2}  \nonumber \\
{\rm subject \ to} &\ g =  \nabla_{x} \ell(P(y|x), f(x; w)) 
\end{align} where $P(y|x)$ is the ground truth label for sample $x$. A natural extension of this idea to SSL is introduced by \cite{miyato2015distributional,miyato2017virtual}. Since SSL algorithms do not have access to ground truth labels $P(y|x)$, their adversarial noise attempts to maximize $\ell(f(x; w), f(x+\Delta x; w))$. But, in the SSL case, the first-order approximation is not useful, because the first derivative of $\ell(f(x; w), f(x+\Delta x; w))$ is always zero at $\Delta x = 0$. Hence, \cite{miyato2015distributional,miyato2017virtual} make a second-order approximation for $\Delta x$ and propose the following approximation to the adversarial noise for each input $x$: \begin{align}
& \Delta x \approx \epsilon_x \frac{g}{||g||_2} \nonumber \\
{\rm subject \ to} &\ g =  \nabla_{\Delta x} \ell( f(x; w), f(x+\Delta x; w) ) \Big|_{\Delta x = \xi d}
\label{vat_eq}
\end{align} where $d \sim N(0,1)$. Therefore, the regularization loss of \cite{miyato2015distributional,miyato2017virtual} is
\begin{align}
& \ell_{VAT}(x; w) := \ell(f(x; w), f(x+\epsilon_x \frac{g}{||g||_2}; w) ) \nonumber  \\
{\rm subject \ to} &\ g =  \nabla_{\Delta x} \ell( f(x; w), f(x+\Delta x; w) ) \Big|_{\Delta x = \xi d}
\label{vat_loss}
\end{align} for one input sample $x$. We will minimize this regularizer as a way of doing input smoothing in our final SSL algorithm. 

\subsection{Weight smoothing} 

Just like the input smoothing, weight smoothing can be formulated as a min-max problem.  More explicitly, \begin{align}
\min_w  &\ \sum_{x_i \in X} \ell(f(x_i; w), f(x_i; w+\Delta w)) \nonumber \\
{\rm subject \ to} &\ \Delta w = \argmax_{||\Delta w||<\epsilon_w} \sum_{x_i \in X} \ell(f(x_i; w), f(x_i; w+\Delta w)) 
\label{opt_weight}
\end{align} Unlike input smoothing, the parameters of both minimization and maximization {\em are the same},  namely the weights of the network, $w$. It is important to note that maximum is taken within a ball of small radius $\epsilon_w$. This appears counter-intuitive at first.

Just like input smoothing can be associated with insensitivity to nuisance variability in the data (hence bias the representation towards invariance), weight smoothing can be associated with generalization, as it has been observed empirically that so-called ``flat-minima''
\citep{hochreiter1997flat,jastrzkebski2017three} corresponds to solutions that tend to yield better generalization. Recent methods \citep{chaudhari2016entropy} try to bias solutions towards such flat minima, including using a conservative penalty \citep{li2014efficient}: 
\begin{align} 
w_t = \argmin_{w} \ell(P(y|x); f(x;w)) + \gamma ||w-w_{t-1}||^2_2.
\end{align} We follow a related, but different, approach: We explicitly find adversarial directions {\em with respect to random subsets of the weights,} then force the network to be robust against these directed perturbation using the remaining weights. 

This problem of optimizing for the worst case is also studied in the robust optimization literature \citep{bertsimas2010robust}. Given “all” the possible adversarial $\Delta w$ values maximizing the loss, they suggest an optimization framework for finding descent directions with second-order cone programming (SOCP) which would guarantee an optimal solution for a convex objective. Since finding all possible adversarial perturbations is not feasible, they find the ones around a ball with gradient ascent and solve SOCP for local solutions iteratively, which is related to our method.  

\subsection{Input smoothing and weight smoothing do not imply each other}
\begin{figure}
  \centering
\includegraphics[width=6cm]{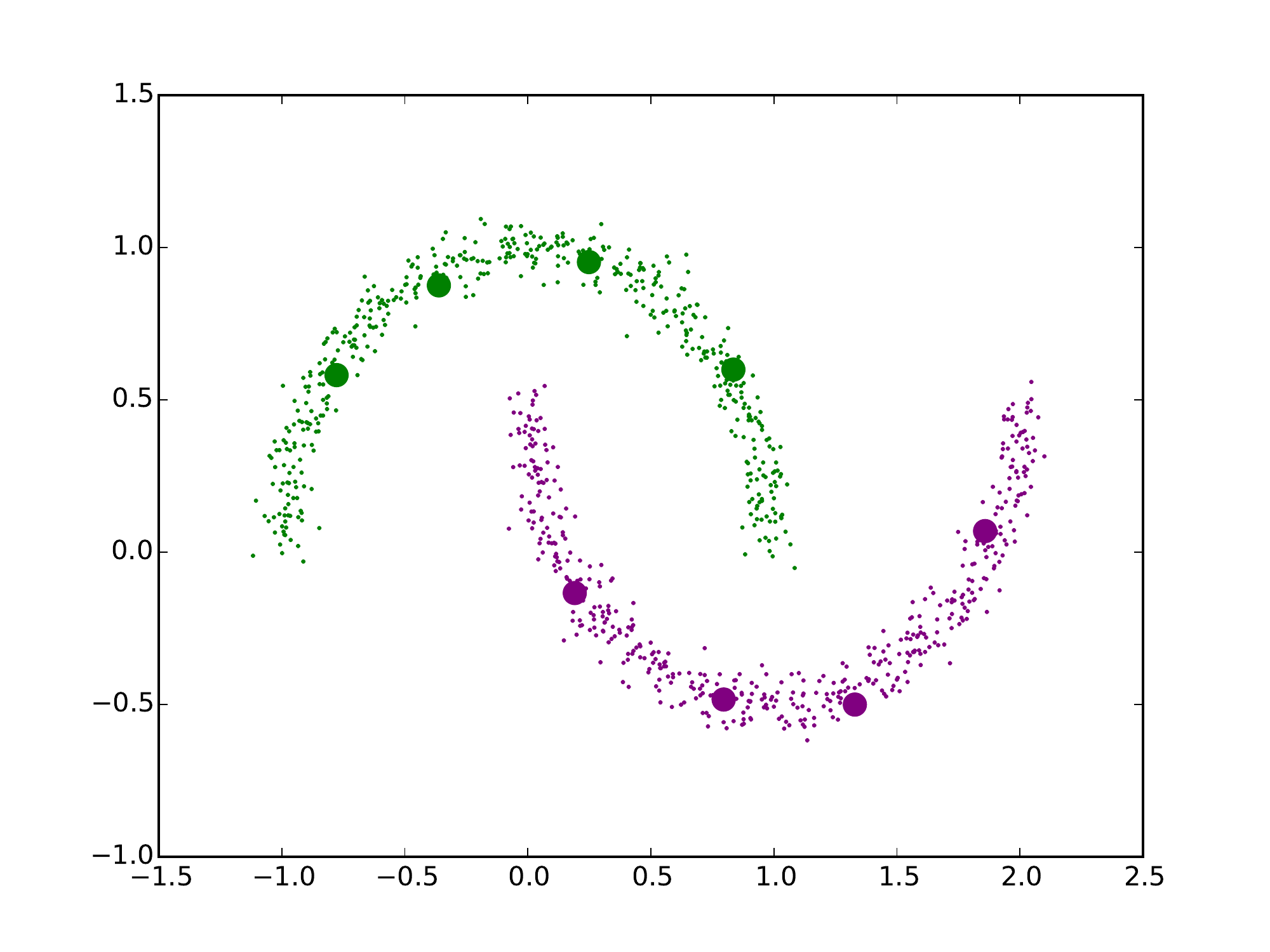}
\includegraphics[width=6cm]{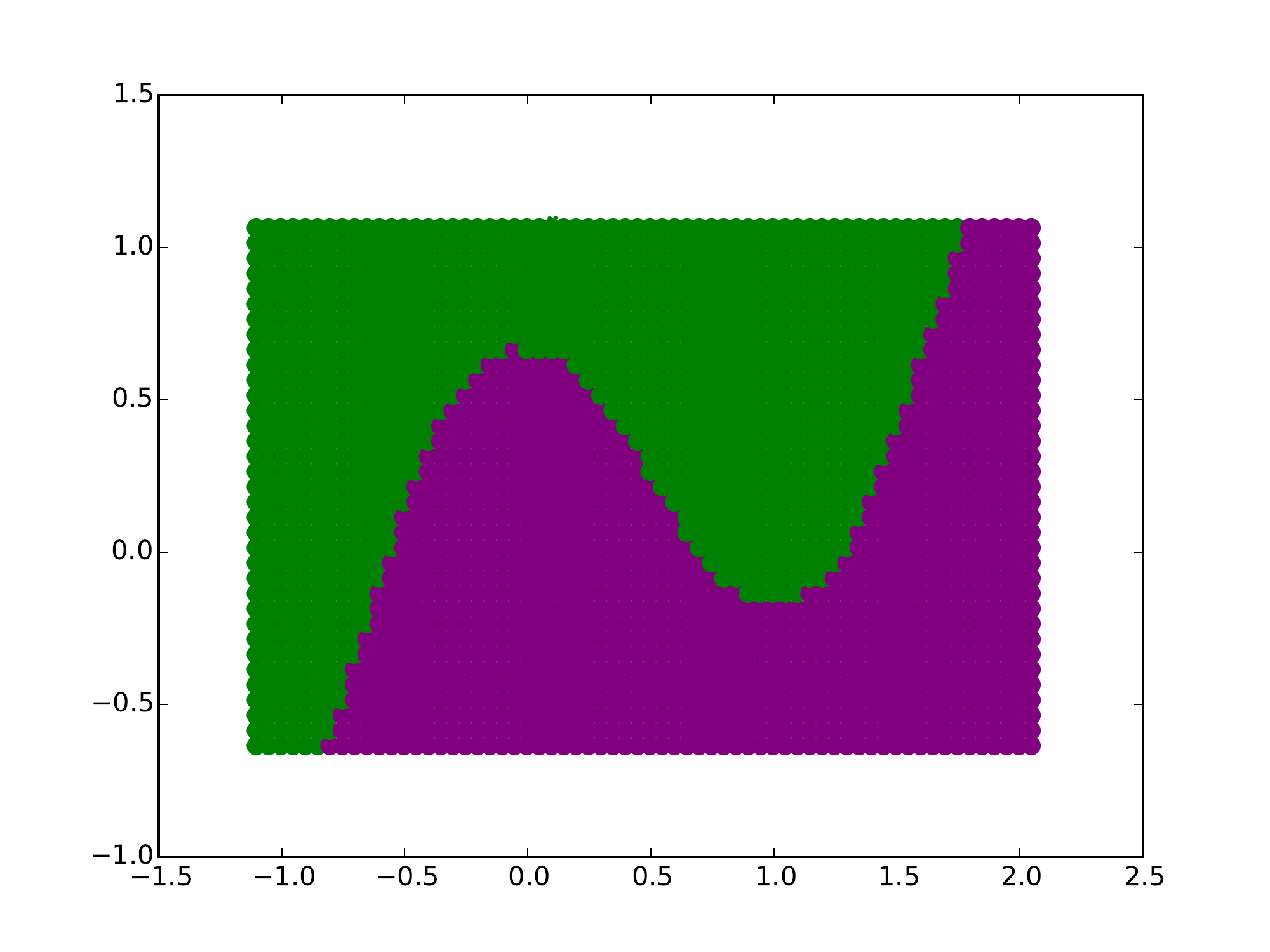}
\includegraphics[width=6cm]{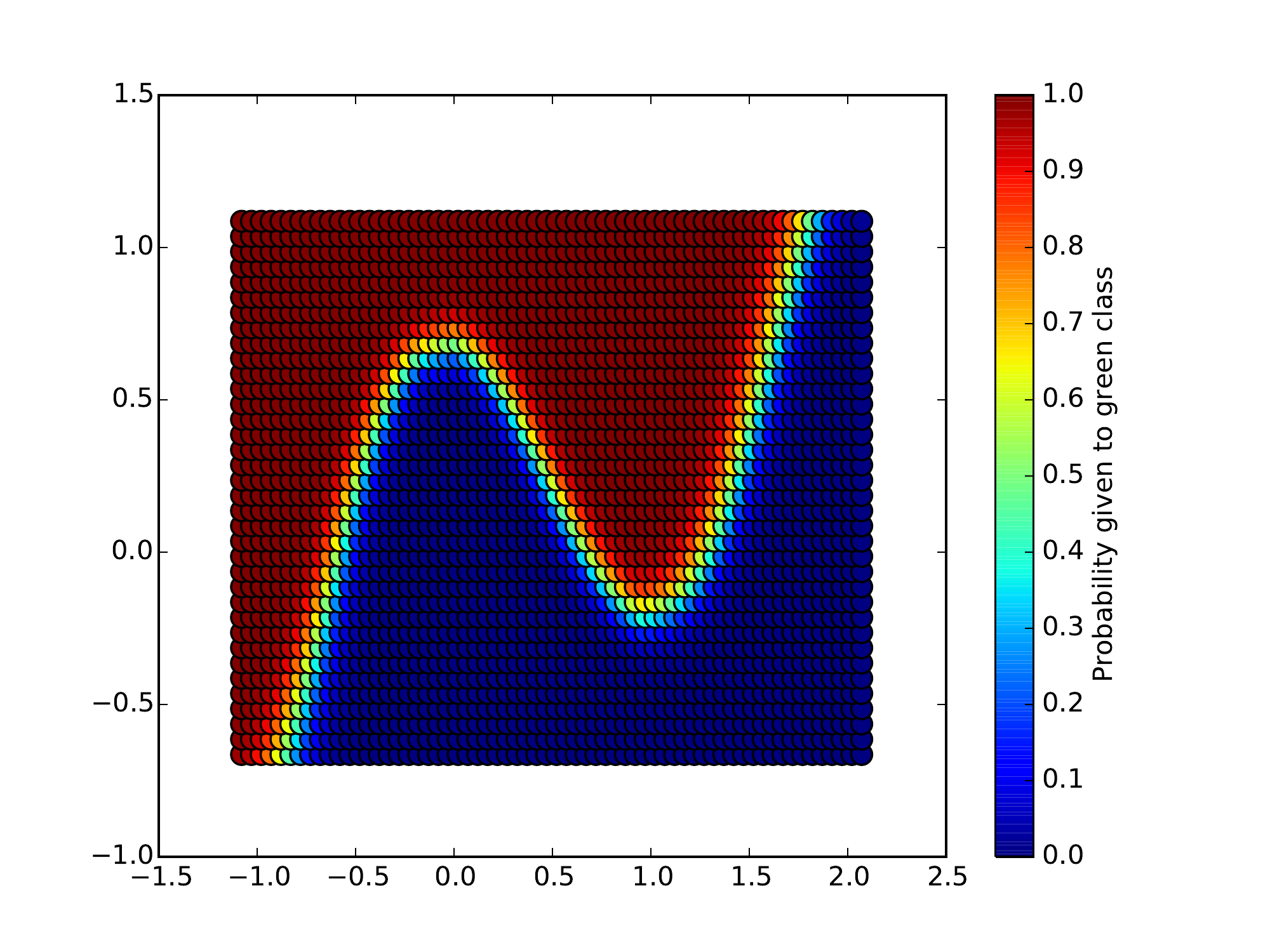}
\includegraphics[width=6cm]{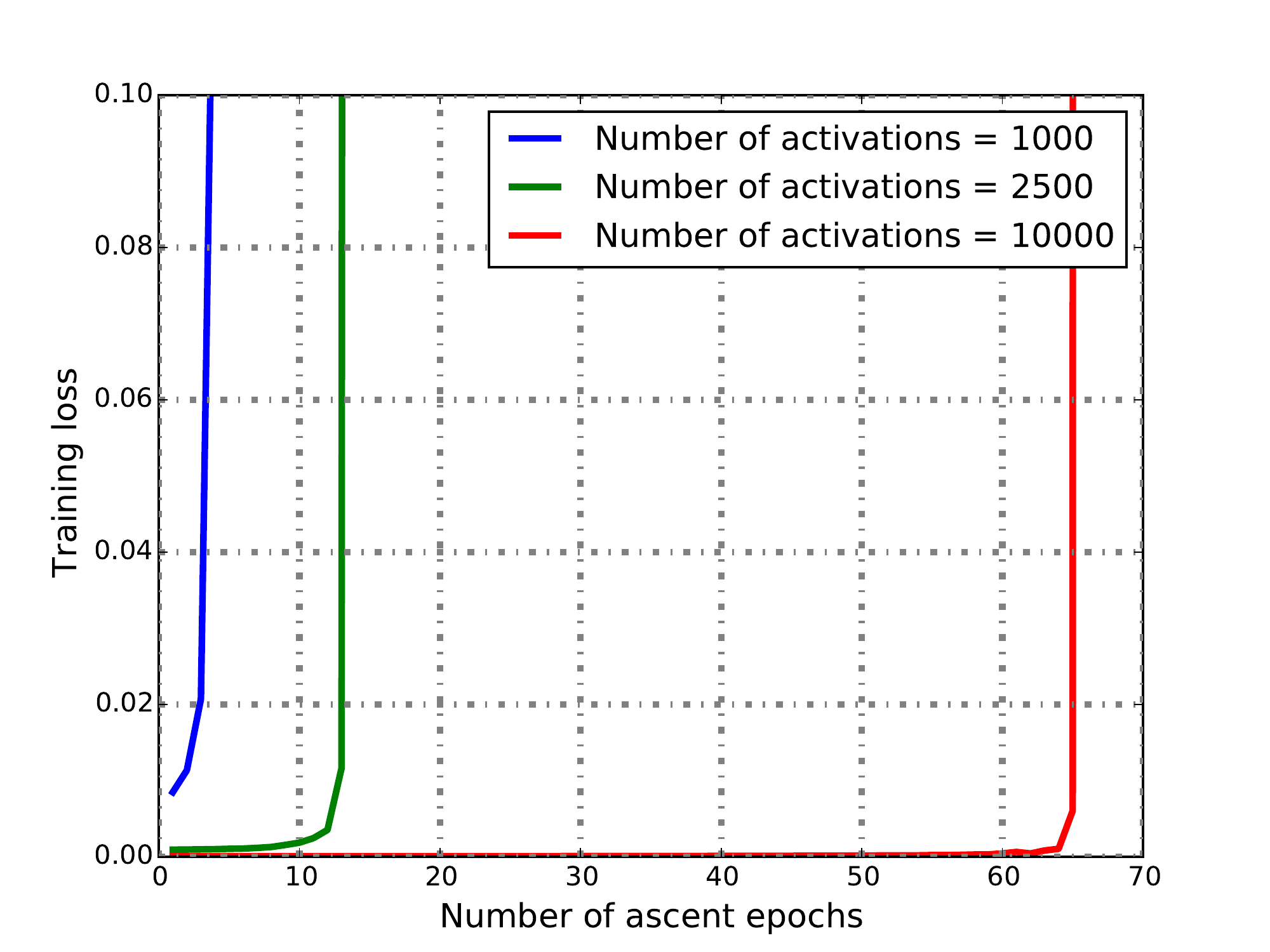}
\caption{\textbf{Over-parametrized networks are more robust to adversarial noises in the weight space even when they have the same decision boundary (i.e. the same input smoothness).} Three MLP networks with different number of hidden units trained with VAT on the half-moon dataset (first panel) have the same decision boundary (second panel). Moreover, the network response (probability given to one of the classes) is almost the identical for each of three networks (third panel). Therefore, robustness of the networks to input perturbation is not just same in ``error" sense but also same in ``loss" sense.  However, the larger the network gets, the more robust it is to the adversarial perturbations in the weight space. Fourth panel shows the training loss versus number of gradient ascent directions for varying sized networks. As it can be seen, having a visible increase in the loss takes more ascent steps for larger networks. This experiment illustrates that networks with the same robustness to input perturbations may have completely different sensitivities to perturbations in the weight space.}
\label{over_param_moon}
\end{figure}

For a linear classifier, the Hessian of the mean-square error (MSE) loss is the data covariance matrix. The local geometry of the loss landscape around the solution where the weights converged does not depend on the classifier structure, nor on its parametrization. The number of zero eigenvalues is determined by the dimensionality of the input data matrix alone. However, it can be easily shown that this is not necessarily the case for nonlinear classifiers. For instance, \cite{sagun2017empirical} shows that the number of zero eigenvalues of the approximate Hessian of the loss has a lower bound of $|w|-N$, where $N$ is the number of training samples and $|w|$ is the number of weights. This simple Hessian argument suggests that the robustness of a network to weight perturbations depends on factors other than its robustness to input perturbations, like the number of parameters in the network, for nonlinear classifiers. Even though they also have empirical evidence for this claim, the spectrum of the Hessian alone is not necessarily a good measure of flatness \citep{dinh2017sharp}. So, we construct a toy experiment to verify that  input smoothness does not necessarily imply  weight smoothness.

We use a half-moon toy dataset whereby there are $4$ labeled (larger circles) and $1000$ unlabeled samples from each cluster (first panel of Fig. \ref{over_param_moon}). We run the VAT algorithm to find the max-margin decision boundary (second panel of Fig. \ref{over_param_moon}). We repeat this experiment for 3 multi-layer perceptron (MLP) networks each having the same structure, except for a different number of weights. The decision boundary is the same, as it can be seen in the second panel of Fig. \ref{over_param_moon}, where they show as one as they overlap perfectly. This implies that for any perturbation in the input space, the increase in the error would be the same for each of these classifiers. To show that this also the case for the loss, we also provide the network response and again they are indistinguishable (third panel of Fig. \ref{over_param_moon}). Hence, input smoothness is the same for each of three MLPs. 

After training three MLP networks with VAT, we apply gradient ascent on the converged weights with a small learning rate to evaluate the robustness of the networks with different number of weights. As it can be seen in the right panel of Fig. \ref{over_param_moon}, it takes more  ascent updates for large networks to diverge. Thus, over-parametrized networks are more robust to perturbations in weight space. This experiment shows that input smoothing does not necessarily imply weight smoothing. I.e. there are factors other than input smoothness determining the geometry of the loss around the converged weight. Another observation is that losses diverge suddenly, implying that it takes several iterations to increase the loss, during which time the landscape is almost constant, before the loss increases sharply. \footnote{This plot implies that for very large networks and small dimensional inputs, reaching to a considerably high loss level set may require many ascent iterations. This might seem counter intuitive when we consider a typical Hessian histogram of a deep network weights as there are usually few large positive eigenvalues. But, it is important to note that the MLP networks we use in this experiment have $4$ hidden layers and the largest of them has $10000$ units per layer. So, the size of network we use for this experiment is much larger than those for which Hessian histograms can be calculated. Unfortunately, calculating the Hessian of such a large network is very expensive computationally. Similarly, it can take many descent iterations to reach a low level loss if the data is too complex for the model. For instance, when training with random labels, it takes many epochs to reduce the loss level \citep{cicek2018saas} even though the loss goes to zero eventually \citep{zhang2016understanding}.} The ascent learning rate chosen in the experiment is $0.005$. During ascent, SGD without momentum and weight decay is used. MLPs are in the form  $2 \rightarrow FC(n) \rightarrow FC(n) \rightarrow FC(n) \rightarrow FC(n) \rightarrow 2$ where $FC(n)$ is fully connected layer followed by a RELU and $n$ is number of hidden units. We report results for $n=1000$, $n=2500$ and $n=10000$.  

Note that we choose to make our experiments in two dimensional inputs to visualize the decision boundary easily and verify that they are the same for different classifiers; not because these results are restricted to small dimensional inputs. These results align with those of \cite{freeman2016topology} where they suggest that the amount of uphill climbing for connecting arbitrary weight pairs is correlated to the size of the network. For more discussion on the effect of over-parametrization on the loss landscape of DNNs see \cite{safran2016quality,sagun2014explorations, baity2018comparing,soudry2016no}.

\subsection{Joint input and weight smoothing}
Once established that input smoothness and weight smoothness are not equivalent, it is natural to try enforcing both. So, the additional regularization we want to have can be framed as follows: \footnote{Note that here we take each smoothness term separately. I.e. adversarial perturbations in weight space $\Delta w$ are calculated for the original images $x$; not for perturbed images $x+\Delta x$. Even though it is possible to formulate the problem such that perturbations in the weight space ($\Delta w$) are functions of perturbations in the input space ($\Delta x$), problem would be even more complicated in that case.} \begin{align}
\min_w  &\ \sum_{x_i \in X} \ell(f(x_i; w), f(x_i; w+\Delta w)) + \ell(f(x_i; w), f(x_i+\Delta x_i; w) )  \nonumber \\
{\rm subject \ to} &\ \Delta w = \argmax_{||\Delta w||<\epsilon_w} \sum_{x_i \in X} \ell(f(x_i; w), f(x_i; w+\Delta w)) \nonumber \\
 &\ \Delta x_i = \argmax_{||\Delta x_i||<\epsilon_x} \ell(f(x_i; w), f(x_i+\Delta x_i; w)) \, \forall x_i \in X
\end{align} 

So, the regularization term we want to minimize for one input $x$ is \begin{align}
L(x; w)  = \ell(f(x; w), f(x; w+\argmax_{||\Delta w||<\epsilon_w} \sum_{x_i \in X} \ell(f(x_i; w), f(x_i; w+\Delta w)))) + \nonumber \\
\ell(f(x; w), f(x+\argmax_{||\Delta x||<\epsilon_x} \ell(f(x; w), f(x+\Delta x; w)); w) ) 
\end{align} 

Then, the overall problem we want to solve in the supervised learning setting becomes
\begin{align}
\min_w  &\  \sum_{x} \ell_{CE}(x; w) + \lambda L(x; w)
\end{align} where \begin{equation}
\ell_{CE}(x; w) = - \langle P(y|x), \log f(x; w)\rangle
\label{ce_loss}
\end{equation} is the cross entropy loss calculated for label estimates $f(x; w)$ and ground truth labels $P(y|x)$. 

Calculating the exact $\Delta x$ for each input is not trivial. Instead, we will use VAT in Eq. \ref{vat_eq} to make our classifier robust against input space perturbations. For achieving weight space smoothness, we will use the proposed algorithm described next. 


\section{
Adversarial Block Coordinate Descent (ABCD)}

\begin{algorithm}
\caption{Adversarial Block Coordinate Descent (ABCD)}
\begin{algorithmic}[1]
  \State Input: Minibatch set $B_t$, loss function $\ell(\cdot)$, initial weights $w_0$.
  \State Hyper-parameters: Ascent and descent learning rates $\eta_{A}$ and $\eta_{D}$. Number of inner iterations $L$.
  \State Output: Final weights $w_L$.
  \For{$l = 1 : L$}
  \State $\Gamma_i$ sample from $\{ 0, -1\}$ for all $i \in \{1,\ldots, |w_0|\}$. 
  \State $\Gamma^a_i = \Gamma_i$ for all $i \in \{ 1,\ldots, |w_0| \}$.
  \State $\Gamma^d_i = \Gamma_i+1$ for all $i \in \{ 1,\ldots, |w_0| \}$.
  \State // Run stochastic gradient \textit{ascent} with a \textit{small} learning rate $\eta_{A}$
  \State $w_{l-\frac{1}{2}} = w_{l-1} - \eta_{A} \Gamma^a \odot \nabla_{w_{l-1}} \left(\frac{1}{|B_t|} \sum_{i = 1}^{|B_t|}  \ell(x_i; w_{l-1}) \right)$
  \State // Run stochastic gradient \textit{descent} with a learning rate $\eta_{D} \gg \eta_{A}$
  \State $w_{l} = w_{l-\frac{1}{2}} - \eta_{D} \Gamma^d \odot \nabla_{w_{l-\frac{1}{2}}} \left(\frac{1}{|B_t|} \sum_{i = 1}^{|B_t|} \ell(x_i; w_{l-\frac{1}{2}}) \right)$ 
  \EndFor
\end{algorithmic}
\label{ACD_pseudo}
\end{algorithm}

Since the parameters of both minimization and maximization are the weights of the network for the min-max problem in Eq. \eqref{opt_weight}, we use a subset of the weights $w$ for finding adversarial directions in weight space and the rest to impose robust to such additive adversarial perturbations in weight space. For this, at each update, we randomly choose half of the weights and take a gradient {\em ascent} step along them with a {\em small} learning rate. Then, on the same batch, we apply gradient descent for the remaining weights with ordinary (larger) learning rate. We call this algorithm Adversarial Block Coordinate Descent (ABCD), for pseudo-code, see Alg. \ref{ACD_pseudo}. ABCD can be considered as an extension of Dropout \citep{hinton2012improving,srivastava2014dropout} and coordinate descent \citep{wright2015coordinate}. However, unlike Dropout, we explicitly regularize our network to be robust against ``adversarial directions" in the weight space; instead of just being robust to zeroed out weights.
\begin{algorithm}
\caption{SSL algorithm using ABCD as optimizer; VAT and entropy as regularizers. $\ell_{CE}(x; w)$, $\ell_{E}(x; w)$, $\ell_{VAT}(x; w)$ are as defined in Eq. \ref{ce_loss}, Eq. \ref{ent_loss}, Eq. \ref{vat_loss} respectively.}
\begin{algorithmic}[1]
  \For{$t = 1 : T$}   
  \State // Run ABCD on cross entropy for labeled samples: 
  \State Sample $B^l_t$
  \State $w_{t'} = ABCD(B^l_t, \ell_{CE}(x; w), w_{t-1})$ // weight smoothing
  \State // Run ABCD on entropy and SGD on VAT loss for unlabeled samples: 
  \State Sample $B^u_t$
  \State $w_{t'} = ABCD(B^u_t, \ell_{E}(x; w), w_{t'})$ // weight smoothing
  \State $w_{t} = SGD(B^u_t, \ell_{VAT}(x; w), w_{t'})$ // input smoothing
  \EndFor    
\end{algorithmic}
\label{ssl_pseudo}
\end{algorithm}

The randomness in ABCD are due to mini-batch optimization that we have to use for computational reasons and the randomness in the mask $\Gamma$ selection. If we ignore these, it would be easier to see the loss minimized by ABCD. The loss minimized by descent in ABCD is the ``worst case'' of the nominal loss landscape. By worst case we mean that, at each point in the weight space, the loss ABCD minimizes is the maximum that can be reached from that point with one small ascent step. In other words, ABCD minimizes the maximum loss around a small ball at each point. Moreover, we do not want the last update for any of the weight parameter to be ascent. So, we do not apply ABCD in the last few epochs.

ABCD can be used for SSL in place of vanilla SGD. We use ABCD for SSL by minimizing the empirical cross entropy for labeled data and the entropy of empirical estimates for the unlabeled data with ABCD. That is, 
\begin{equation}
\ell_{E}(x; w) = - \langle f(x; w), \log f(x; w)\rangle
\label{ent_loss}
\end{equation} for unlabeled data which is a well-known regularizer in the SSL literature \citep{dai2017good,grandvalet2005semi,krause2010discriminative,springenberg2015unsupervised}. 

We additionally report performance of ABCD combined with VAT to see the effect of applying both input and weight smoothing. The pseudo-code using VAT as loss function and ABCD as optimizer for SSL task is given in Alg. \ref{ssl_pseudo}. For labeled data, ABCD used as an optimizer to minimize cross entropy to achieve weight smoothing. We do not minimize $\ell_{VAT}(x;w)$ for labeled data as proposed in the corresponding paper. For unlabeled data, ABCD is used to minimize entropy $\ell_{E}(x;w)$ and SGD is used to minimize $\ell_{VAT}(x;w)$ for weight and input smoothing respectively. We set $\eta_A = 10^{-5}$ for all of our experiments. $\eta_D$ can be chosen as usual with high initial value from $\{0.1, 0.01\}$ and is decreased during training. In all the SSL experiments, we only use the network called conv-large from \cite{miyato2017virtual,tarvainen2017mean}. Only translation and horizontal flipping are used as data augmentations to allow fair comparison with some of the previous SSL algorithms. Horizontal flipping is only used in CIFAR10.

In our implementation of VAT \citep{miyato2017virtual}, we set  $\epsilon_x$ in Eq. \ref{vat_eq} to be $128$ for CIFAR10 and $0.25$ for SVHN. Even though we search $\epsilon_x$ in a fine grid, we could not get to the performance reported in their paper for CIFAR10; possibly because we use a different optimizer (SGD instead of ADAM), different learning rate scheme and different ZCA regularizer. The performance of our VAT implementation is given in Table \ref{ssl_soa}. Our implementation ($13.28\%$) of VAT without entropy minimization is about $2$ percent worse than what is reported in \cite{miyato2017virtual} ($11.36\%$) for CIFAR10. But, we still use the numbers reported in \cite{miyato2017virtual} for comparison purposes in Table \ref{ssl_soa}. Additional details on training given in the Supplementary Material.

\section{Evaluation}

\begin{figure}
  \centering
\includegraphics[width=6cm]{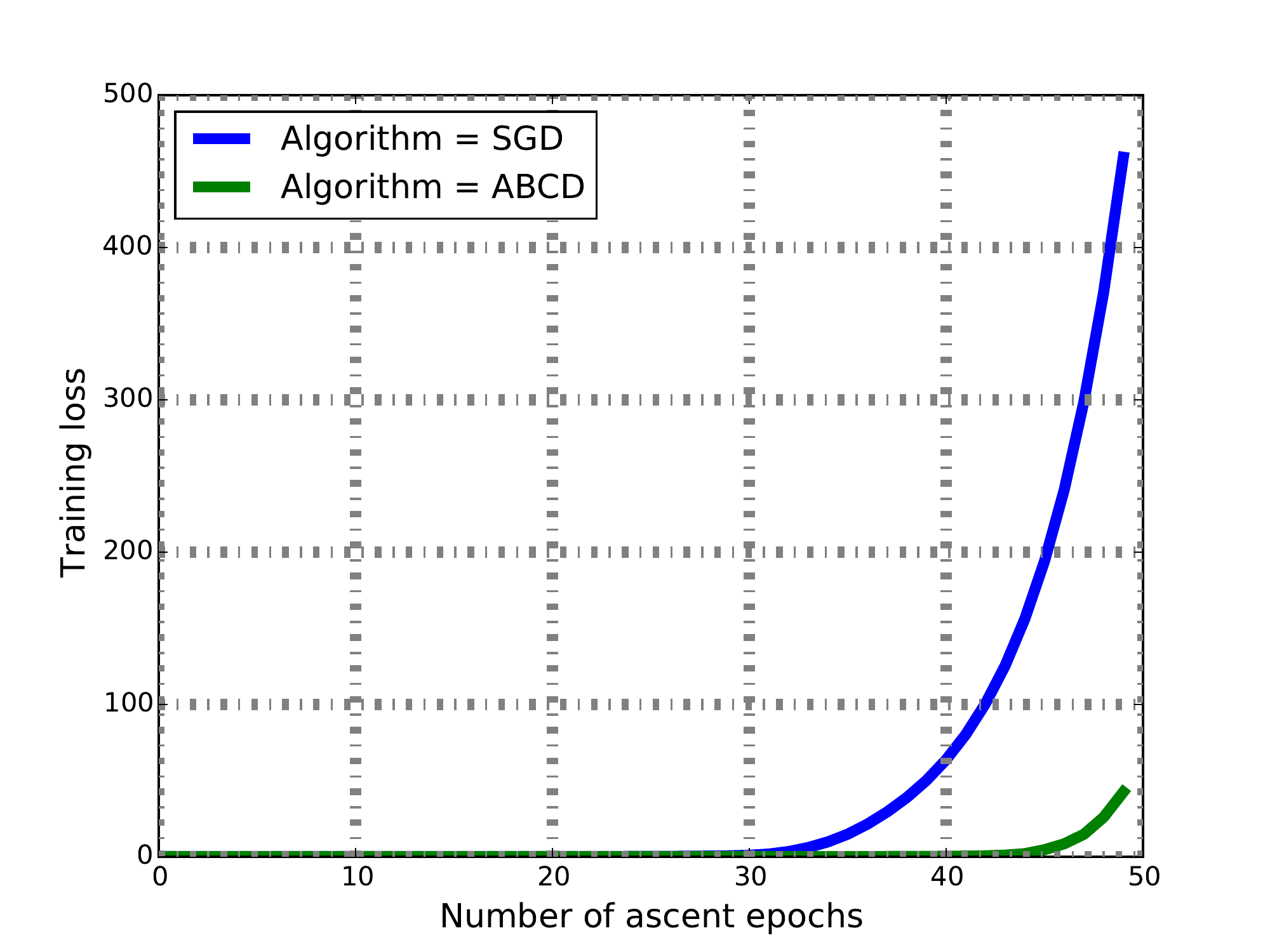}
\includegraphics[width=6cm]{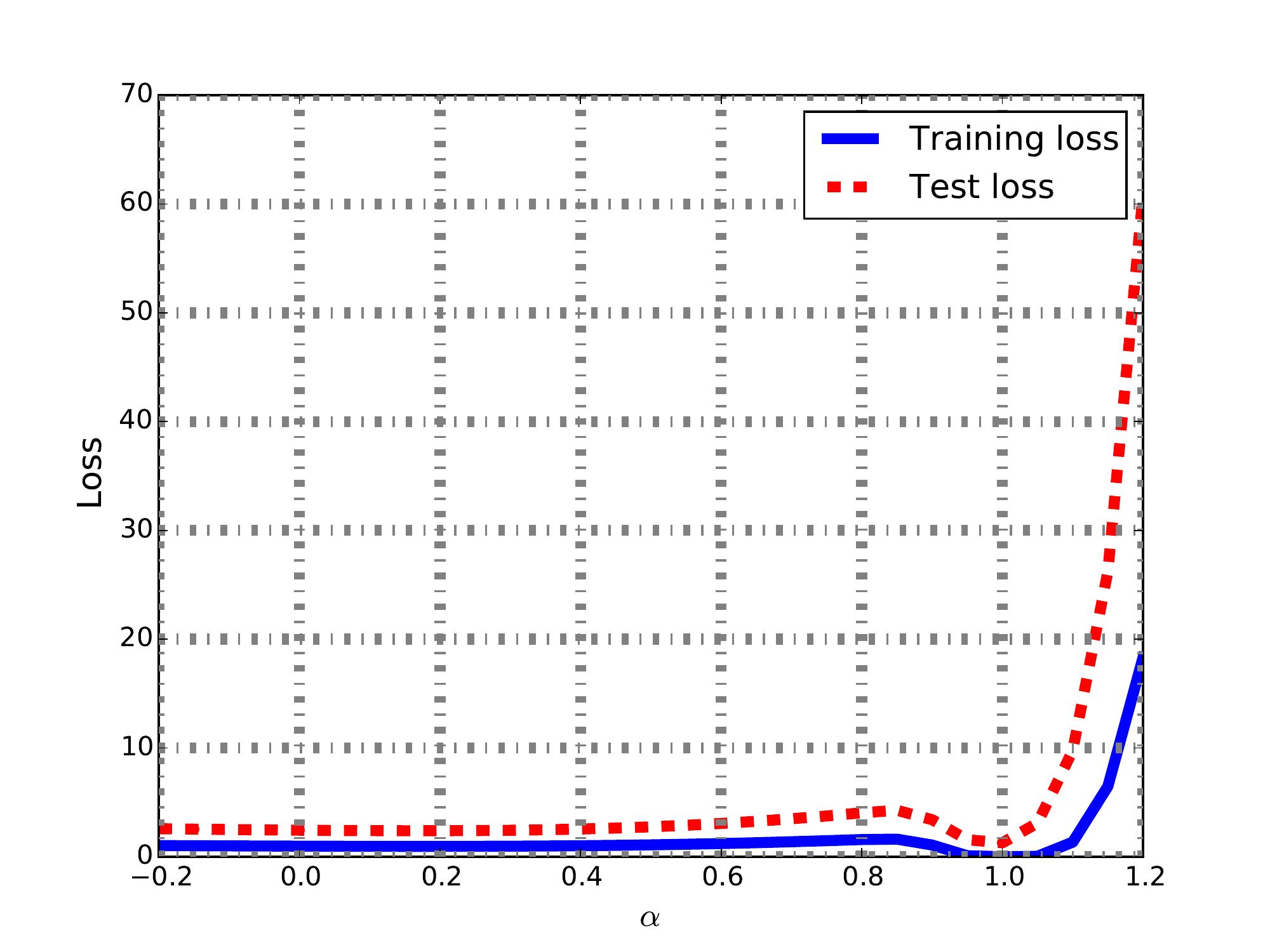}
\caption{\textbf{Robustness of ABCD vs SGD trained networks to ascent updates}. Starting from the ABCD and SGD trained weights, we apply gradient ascent with learning rate $0.001$ on the cross entropy loss. This plot shows the increase in the training loss versus number of ascent steps. Both weights diverge quickly, but ABCD trained network diverges later than that of SGD verifying the robustness of ABCD to weight space perturbations. \textbf{1D visualization of loss landscapes of SGD and ABCD trained weights}. Training and test losses over the curve $\alpha*w_{SGD} + (1-\alpha)*w_{ABCD}$ are given as suggested by \cite{goodfellow2014qualitatively} where $\alpha \in [-0.2, 1.2]$. $\alpha=0$ and $\alpha=1$ correspond to the weights of ABCD and SGD trained networks respectively. This method compares the flatness of two methods in one direction only and in that direction, ABCD seems much more robust to weight perturbations.}
\label{1D_visualization}
\end{figure}

\begin{table}
\begin{center}
 \begin{tabular}{||c  c  c||} 
 \hline
  \multirow{3}{*}{SSL Method} & \multicolumn{2}{c||}{Test error rate (\%)}  \\ [0.5ex]
& CIFAR10 & SVHN \\ [0.5ex]
& $N^l = 4,000$ & $N^l = 1,000$ \\ [0.5ex]
& $N^u = 46,000$ & $N^u = 72,257$ \\ [0.5ex]
\hline\hline
VAT+EntMin \cite{miyato2017virtual} & $10.55$ &  $3.86$ \\ 
\hline
Stochastic Transformation \cite{sajjadi2016regularization} & $11.29$ & NR \\
\hline
Temporal Ensemble \cite{laine2016temporal} & $12.16$ & $4.42$  \\
\hline
GAN+FM \cite{salimans2016improved} & $15.59$ & $5.88$  \\
\hline
Mean Teacher \cite{tarvainen2017mean} & $12.31$ & $3.95$  \\
\hline\hline
EntMin (our implementation) & $15.42$ $\pm$ $0.22$ & $6.02$ $\pm$ $0.02$\\
 \hline
 VAT without EntMin (our implementation) & $13.28$ $\pm$ $0.11$ &  $5.60$ $\pm$ $0.29$\\
\hline
 VAT+EntMin (our implementation) & $11.81$ $\pm$ $0.07$ & $4.10$ $\pm$ $0.06$\\
 \hline\hline
 ABCD+EntMin & $11.98$ $\pm$ $0.04$ & $4.93$ $\pm$ $0.04$ \\
 \hline
 ABCD+EntMin+VAT & $10.11$ $\pm$ $0.14$ & $3.63$ $\pm$ $0.03$	 \\
 \hline
\end{tabular}
\caption{{\bf Comparison with the state-of-the-art on SSL tasks.} Error rates on the test set are given for CIFAR10 and SVHN. NR stands for ``not reported.'' CIFAR10 is trained using 4,000 labeled and 46,000 unlabeled samples, SVHN using 1,000 labeled and 72,257 unlabeled samples. Results are averaged over three random labeled sets. We report performance of ABCD alone and combined with VAT. ABCD+EntMin+VAT refers to algorithm in Alg. \ref{ssl_pseudo} where ABCD is used as an optimizer; entropy and VAT are used as regularizers in the loss function.  ABCD+EntMin uses only entropy for unlabeled data to report performance of ABCD without VAT. {\bf SSL baselines.} Baseline algorithms are EntMin, VAT and VAT+EntMin. EntMin minimizes the entropy of estimates for unlabeled data with standard SGD. Similarly, VAT minimizes $\ell_{VAT}$ from Eq. \ref{vat_loss} and VAT+EntMin minimizes both on unlabeled data. Note that these are the results we get with our own implementation of VAT. }
\label{ssl_soa}
\end{center}
\end{table}
To evaluate the robustness of ABCD-trained weights to adversarial weight space perturbations, we report the number of ascent updates necessary for the loss to diverge. In Fig. \ref{1D_visualization} (left), the plot of training loss v.s. number of ascent updates is given for SGD and ABCD-trained weights. As it can be seen, the number of ascent updates needed for the loss to diverge for ABCD  is more than for SGD. 

Another way of comparing the local geometrical properties of the weights is by visualizing the loss landscape around the converged weights. As the deep networks we use have  very high dimension, several visualization tricks have been suggested to visualize the loss landscapes in $1$- or $2$-dimensional subsets of the weight space \citep{li2017visualizing,goodfellow2014qualitatively,keskar2016large}. We employ the technique suggested in \cite{goodfellow2014qualitatively} in Fig. \ref{1D_visualization} (right): we plot the loss on the curve $\alpha*w_{SGD} + (1-\alpha)*w_{ABCD}$ where $\alpha \in [-0.2, 1.2]$. $w_{SGD}$ and $w_{ABCD}$ are the weights converged when training with SGD and ABCD respectively. As it can be seen, the loss does not increase around ABCD trained weights. We also report Hessian histograms in the Supplementary Material. Further details on the networks and datasets used in the experiments given there.

In Table \ref{ssl_soa}, we compare the performance of ABCD with state-of-the-art SSL algorithms. We report performance of the proposed algorithm on SVHN \citep{netzer2011reading} and CIFAR10 \citep{krizhevsky2009learning} in SSL setting. SVHN consists of $32 \times 32$ images of house numbers. We use $73,257$ samples for training, rather than the entire $600,000$ images; $26,032$ images are separated for evaluation. CIFAR10 has $60,000$ $32 \times 32$ images, of which $50,000$ are used for training and $10,000$ for testing. We choose labeled samples randomly. We also choose them to be uniform over the classes as it is done in previous works \citep{miyato2017virtual}. In CIFAR10, $4,000$ and in SVHN $1,000$ of training samples are labeled. Except \cite{sajjadi2016regularization}, all the methods use modest augmentations (translation and horizontal flipping) and do not exploit recent deep learning models like ResNet \citep{he2016deep}. We report ABCD with entropy minimization alone and combined with VAT. When we run ABCD only with entropy minimization, it yields second-best SSL performance after VAT in CIFAR10 excluding \cite{sajjadi2016regularization}. Combining ABCD with VAT improves the scores verifying that they are complementary.

\section{Discussion and Related Work}
\label{sec:discussion}

Next, we discuss our contribution in the context of related and recent work in SSL. 

\textbf{Input smoothing in SSL.} In addition to work referenced earlier, there are graph-based methods \citep{solomon2014wasserstein,yang2016revisiting} that penalize having different labels for similar input pairs. For instance, given that $s_{ij}$ is the measure of similarity for input samples $x_i$ and $x_j$, they minimize the energy $s_{ij} (f(x_i; w) - f(x_j; w))^2$ performing label propagation under the constraint of fitting to labeled data. This forces the discriminant to change little in response to different inputs with large $s_{ij}$. Generative models by \cite{springenberg2015unsupervised} suggests that they maximize the margin with the help of fake samples.  

\textbf{Weight smoothing in SSL.} Teacher-student methods \citep{laine2016temporal,tarvainen2017mean} average over many predictions or weights in a way that the teacher network can attract student networks towards itself. A similar algorithm is suggested by \cite{zhang2015deep} for parallel computing under communication constraint where each replica is attracted to the reference system. \cite{baldassi2016unreasonable} studies such algorithms for models with discrete variables and they argue that they achieve to find robust local minimas. Thus, one can relate the success of teacher-student models of state-of-the-art deep SSL algorithms for their ability to converge to robust weights. We recently became aware that \cite{park2017adversarial} also combine VAT with Virtual Adversarial Dropout (VAdD) and, like us, improve upon the VAT baseline as a result.  VAdD finds a zero mask of dropout adversarially at each update rather than trying to be robust against adversarial ``directions". We find the latter approach more effective in achieving weight smoothing and avoiding convergence to locations in the loss landscape with strong ascent directions. As a result, we reach comparable performance to \cite{park2017adversarial} without using sophisticated rate scheduling like ramp-up and mean-only batch-normalization \citep{salimans2016weight}.

\textbf{Effect of noise to generalization.} Adding random noise to gradients is known to improve the generalization \citep{welling2011bayesian}. \cite{jastrzkebski2017three} analyzes the effect of the inherent noise due mini-batch usage in the properties of point converged by SGD. They conclude that for larger noise, network favors wider minima more under the assumption that the noise is isotropic. ABCD differs from these works by adding adversarial noise to weights instead of random noise.

\subsubsection*{Acknowledgments}
SC would like to thank Pratik Chaudhari for fruitful discussions and valuable suggestions.

{
\footnotesize
\bibliographystyle{apalike}
\bibliography{ref}
}

\end{document}